\documentclass[10pt,twocolumn,letterpaper]{article}

\usepackage{wacv}

\usepackage{times}
\usepackage{epsfig}
\usepackage{graphicx}
\usepackage{amsmath}
\usepackage{amssymb}
\usepackage{xcolor}
\usepackage[utf8]{inputenc} % allow utf-8 input
\usepackage[T1]{fontenc}    % use 8-bit T1 fonts
\usepackage{url}            % simple URL typesetting
\usepackage{booktabs,multirow}       % professional-quality tables

\usepackage[accsupp]{axessibility}  % Improves PDF readability for those with disabilities.

\usepackage{amsfonts}       % blackboard math symbols
\usepackage{nicefrac}       % compact symbols for 1/2, etc.
\usepackage{microtype}      % microtypography

\usepackage{makecell} % for bold in table using \small
\usepackage{tabularx, ragged2e} 
\usepackage{booktabs}

\usepackage{graphicx}
\usepackage{comment}
\usepackage{amsmath,amssymb} % define this before the line numbering.
\usepackage{color}

\newcommand{\figref}[1]{Fig.~\ref{#1}}
\newcommand{\tabref}[1]{Tab.~\ref{#1}}
\newcommand{\secref}[1]{Sec.~\ref{#1}}

\newcommand{\equref}[1]{Eq.~(\ref{#1})}

\newcommand{\beas}{\begin{eqnarray*}}
	\newcommand{\eeas}{\end{eqnarray*}}
\newcommand{\bea}{\begin{eqnarray}}
\newcommand{\eea}{\end{eqnarray}}
\newcommand{\bes}{\begin{equation*}}
\newcommand{\ees}{\end{equation*}}
\newcommand{\be}{\begin{equation}}
\newcommand{\ee}{\end{equation}}

\newcommand{\cR}{{\cal R}}

\newcommand{\norm}[1]{\left\lVert#1\right\rVert}

\usepackage{multicol}

\def\eg{\emph{e.g}\onedot} 
\def\ie{\emph{i.e}\onedot} \def\Ie{\emph{I.e}\onedot}
 
 \def\vs{\emph{vs}\onedot}
 
\def\etal{\emph{et al}\onedot}
\def\wacvPaperID{544} % Enter the WACV Paper ID here

\wacvfinalcopy % *** Uncomment this line for the final submission

\ifwacvfinal
\fi

\ifwacvfinal
\usepackage[breaklinks=true,bookmarks=false]{hyperref}
\else
\usepackage[pagebackref=true,breaklinks=true,colorlinks,bookmarks=false]{hyperref}
\fi

\begin{document}

%%%%%%%%% TITLE
\title{Video and Text Matching with Conditioned Embeddings}

\author{Ameen Ali\\
Tel Aviv University\\
%{\tt\small ameenali023@gmail.com}
\and
Idan Schwartz\\
Technion\\
NetApp\\ 

%{\tt\small idansc.github.io}
\and
Tamir Hazan\\
Technion\\
%{\tt\small tamir.hazan@technion.ac.il}
\and
Lior Wolf\\
Tel Aviv University\\
%{\tt\small wolf@cs.tau.ac.il}
}

\maketitle

\ifwacvfinal
\thispagestyle{empty}
\fi

%%%%%%%%% ABSTRACT
\begin{abstract}
We present a method for matching a text sentence from a given corpus to a given video clip and vice versa. Traditionally video and text matching is done by learning a shared embedding space and the encoding of one modality is independent of the other. In this work, we encode the dataset data in a way that takes into account the query's relevant information.  The power of the method is demonstrated to arise from pooling the interaction data between words and frames. Since the encoding of the video clip depends on the sentence compared to it,  the representation needs to be recomputed for each potential match. To this end, we propose an efficient shallow neural network. Its training employs a hierarchical triplet loss that is extendable to paragraph/video matching. The method is simple, provides explainability, and achieves state-of-the-art results for both sentence-clip and video-text by a sizable margin across five different datasets: ActivityNet, DiDeMo, YouCook2, MSR-VTT, and LSMDC. We also show that our conditioned representation can be transferred to video-guided machine translation, where we improved the current results on VATEX. Source code is available at \url{https://github.com/AmeenAli/VideoMatch}.
\end{abstract}

%%%%%%%%% BODY TEXT
\section{Introduction}

The process of matching data from two domains is often seen as a sequential process of comparison between two static representations, obtained by preprocessing these domains. However, it is possible that a more effective way is to consider recent models for cognitive search, in which the content of the query signal from the first domain changes our perception of the data from the second domain.

\begin{figure}[t]
	\includegraphics[width=1\linewidth]{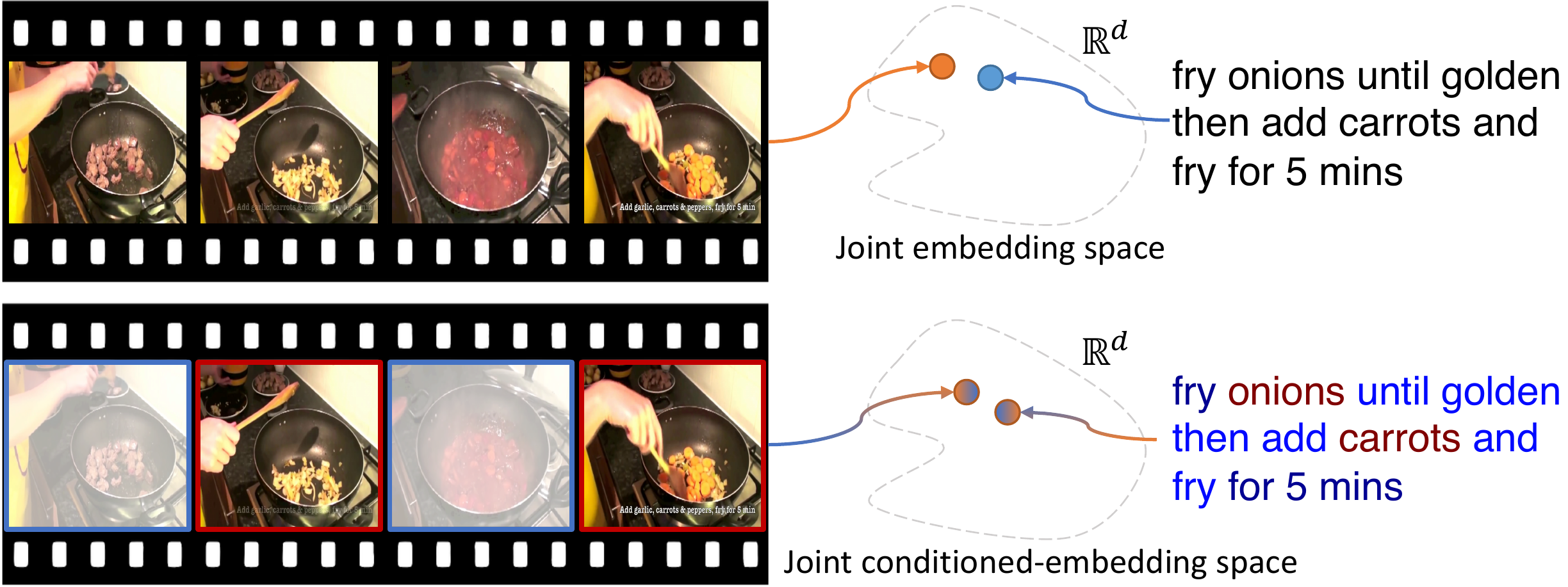}
	%\captionsetup{justification=centering}
	\caption{Current approaches use a joint embedding space for efficient video and text retrieval via a dot-product (illustrated in the top row). However, retrieving the relevant video to a given text query requires a concise representation that consists of only the relevant frames.  For this reason, we propose a new conditioned embedding that encodes all the videos conditioned on a given query. On the bottom row, we highlight with red color the relevant words given the video clip. We also mark with a red border the relevant frames given the text query.  Note the query is ``fry onions ... add carrots ...'', and the selected relevant frames show onion and carrots, while the non-relevant frame shows meat.}
	\label{fig:teas}
\end{figure}

In this work, we consider hierarchical representations of two domains, in which repeated interaction between the two domains modifies the representations of each data source. In particular, we consider the problem of matching multi-clip video to multi-sentence text. The interaction-based representations are a form of attention. As seen in \figref{fig:teas}, when matching the sentence ``fry onions until golden then add carrots and fry for 5 mins'' to a cooking video segment, the frames showing onions, or carrots are highlighted in the video representation.
Focusing on the video and text domains enables us to benefit from the extensive literature on image and text embedding, as well as from the existence of multiple existing benchmarks. 

We represent the video frames with a pre-trained CNN and the words using the GloVe~\cite{pennington2014glove} representation, both of which are standard. Then, a temporal context is added to the video by applying a unidirectional GRU~\cite{cho2014properties}. This is done at the clip level for video and the sentence level for the text. Given a video clip and a possibly-matching sentence, we compute an affinity score between every frame (clip) and word (sentence) by projecting the domain-specific representations to a shared vector space, in which the cosine similarity is employed.

The individual similarities are aggregated to define per frame (per word) weights, which are used to obtain one weighted-sum representation of the video clip (sentence). The video clips and sentences are then each processed by another pair of GRUs to create another layer of representation, whose initial state depends on information collected from the entire video and text paragraph. Max pooling is then employed to obtain a single vector representing each video (all clips) and each paragraph (all sentences). These representations are projected to a joint vector space, in which they are compared.

Since the encoding of the video clip depends on the sentence it was compared to, the representation needs to be recomputed for each potential match. While this does not affect the asymptotic complexity of the search process, it does require applying the encoding mechanism $N^2$ times, instead of $2N$ times, where $N$ is the number of videos. However, in practice, we had no challenge running on the existing benchmarks, using a shortlist obtained using static embeddings.  %  method employs an efficient single layer dot-product mechanism.  Meanwhile, if complexity is an issue for a specific application, one can consider 

 The new method is able to outperform by a sizable margin the state of the art results for the cross-modality text/video retrieval on a diverse set of video to text matching benchmarks, including ActivityNet~\cite{krishna2017dense}, DiDeMo~\cite{anne2017localizing}, YouCook2~\cite{ZhXuCoCVPR18} , Vatex~\cite{wang2019vatex} , MSR-VTT~\cite{xu2016msr} and LSMDC~\cite{rohrbach2015dataset}. The robustness of our method is further demonstrated by its application to the task of video-guided based translation on Vatex, surpassing the recent baseline.

\section{Related Work}
In the task of {\bf video/text retrieval}, given a set of video descriptions,  we rank them according to video relevancy. Early works clustered videos and captions by considering metadata, video tags, and video features(\ie, SIFT, BoW)~\cite{aradhye2009video2text, wei2009multimodal, huang2012multi}. Afterward, a model that finetunes video caption generation model was proposed~\cite{krishna2017dense}.  However, a dedicated matching model significantly improved the results~\cite{zhang2018cross}. We follow the same strategy but improve in two aspects: (i) we condition the video and paragraph embeddings, and (ii) we use a weighted interaction score, instead of a dot product. 
The video-paragraph retrieval relies on dense clip annotations, which are not always available. Therefore, instead, recent methods focus on the sentence-video retrieval variant. Liu~\etal\cite{liu2019use} utilize different experts such as speech, audio, motion, OCR, appearance, and face detection. Gabeur~\etal\cite{gabeur2020multi}, improved this approach by leveraging transformers networks. Our approach is much more straightforward and considers only the video and the text. Lei~\etal\cite{lei2021less} suggested the use of just a few sparsely sampled clips for matching. Patrick~\etal\cite{patrick2020support} propose incorporating a generator and a contrastive loss to consider semantically-related negative samples better. Different from all the methods mentioned above, we are the first to study the power of conditioned video/clip embedding, which results in a significant performance improvement. Recently, the HowTo100M dataset has been used as a basis for training large-scale approaches  \cite{miech2019howto100m, zhu2020actbert, luo2020univl}.  Notably, out approach remains competitive even when compared to pre-trained approaches, despite the significant increase in data pre-training methods use.

%In this work, to fairly compare with most retrieval models, we report the results without large-scale pre-training. 

In \textbf{temporal grounding}, where the task is to localize a segment in the video  that  semantically  corresponds  to  the  given  sentence. Most of the existing approaches to this problem use dense supervision of event locations for training~\cite{anne2017localizing,gao2017tall, xu2019multilevel,chen2019semantic, xu2019multilevel, chen2019semantic, yuan2019find, zhang2019man}. These annotations are not always available and recent methods, such as Mithun~\etal focus on weakly-supervised temporal grounding by leveraging text-to-frame attention scores~\cite{tanLoGAN2019, mithun2019weakly,gao2019wslln,duan2018weakly}. The most relevant to our approach, Yuan~\etal~\cite{yuan2019find} uses co-attention for video localization. However, our co-attention is simpler as it works in a parallel manner, while Yuan~\etal infers attention in an alternating manner, \ie, the attention is computed multiple times sequentially.

{\bf Hierarchical encoding} deals with long text (\ie, paragraph) or a long video with multiple clips. Li~\etal\cite{li2015hierarchical} propose encoding a paragraph with a hierarchical autoencoder. Following, a hierarchical RNN was proposed~\cite{pan2016hierarchical, yu2016video}. Niu~\etal\cite{niu2017hierarchical} leverage the hierarchy of a sentence parse tree to match a sentence with an image. Chen~\etal\cite{chen2020fine} proposed using hierarchical graph encoding on the local and global level of the data. Similar to our approach, Zhang~\etal\cite{zhang2018cross} embed video and paragraph in a two-level hierarchical manner. The first level embeds the clips and sentences, followed by a second level that embeds the video and paragraph. However, the embeddings are independent. In this work, we demonstrate that conditioning the clip embedding on the matched sentence, and vice versa significantly improves the performance. 

{\bf Multimodal Attention} has been a prominent tool for modeling the interaction between the two domains, since it models interactions in a way that selects important frames and words. We follow the same practice and use efficient interaction-based attention as a  tool for conditioning.  In early work,  image attention was proposed for improved image captioning~\cite{xu2015show}. This idea was later extended to visual question answering~\cite{xu2016ask}. Other approaches improved the vector-fusion operator for better interaction modeling~\cite{fukui2016multimodal,kim2016hadamard,ben2017mutan}. Some approaches imitate a multi-step reasoning with stacked attention modules sequentially~\cite{yang2016stacked}.  Lu~\etal~\cite{lu2016hierarchical} proposed a co-attention module that attends to visual and textual modalities jointly.  A more advanced approach for co-attention used a bilinear module that efficiently generates attention for every pair~\cite{kim2018bilinear}. Next, a general factor-based framework that allows joint attention to any number of modalities was proposed~\cite{schwartz2017high, schwartz2019factor}.  In recent work, the factor graph attention is used to allow interaction between video frames or list of images~\cite{schwartz2019simple, braude2021towards}. 

\section{Method}

Learning a similarity measure between text-to-video allows an efficient video retrieval of a video given text. In the following, we propose to learn from training data a similarity score between a video $v \in V $, and a paragraph $p \in P$. The training dataset $ \mathcal{S} = \{ (v_1, p_1), \ldots, (v_n, p_n) \}$ consists of $n$ video samples and matching paragraphs. Each video sample $ \boldsymbol{v_i} \in V $ is a sequence of clips $ \boldsymbol{v_i} = \{ c^i_1, \ldots, c^i_{n_{^i}} \} $ and each clip $j$ is composed of frames $ \boldsymbol{c^i_j} = \{ f^{ij}_1, \ldots, f^{ij}_{n_{{ij}}} \} $. The matching paragraph of a video/paragraph pair $ (\boldsymbol{v_i},\boldsymbol{p_i}) $ consists of a sequence of sentences $ \boldsymbol{p_i} = \{s^i_1, \ldots, s^i_{n_i} \} $, while for any $j = 1,...,n_i$ the sentence $\boldsymbol{s^i_j}$ corresponds to the clip $ \boldsymbol{c^i_j} $. The sentence itself is composed of words $ \boldsymbol{s^i_j} = \{ w^{ij}_1, \ldots, w^{ij}_{m_{ij}} \} $ where $ m_{ij} $ is the number of words in the $j$-th sentence $\boldsymbol{s^i_j}$ of paragraph $\boldsymbol{p_i}$. We note that although the $j$-th clip and the $j$-th sentence, of the a matching video/paragraph pair of any index $i$, are matched, their words and frames do not match, and the number of frames in a clip $n_{ij}$ often differ from the number of words $m_{ij}$ in its corresponding sentence.  

Our training procedure is learning the similarity of paired video and paragraph $(\boldsymbol{\boldsymbol{v_i}},\boldsymbol{p_i}) \in \mathcal{S}$ by embedding both $\boldsymbol{v_i}$ and $\boldsymbol{p_i}$ in a joint space $\cR^{d_{vp}}$. Each such training pair should lie close together in the joint space, while embedding a paragraph and a video that do not match should lie far apart. The embedding of the video and paragraph are dependent on each other. To search for the matching paragraph of a given video, we consider the pairwise Euclidean distances between the embedding of the video given each possible paragraph and the embedding of each paragraph given the video. The same process, in the other direction, is used for searching for the matching video given a paragraph. 

\subsection{Joint Clip and Sentence Encoding} The dataset $ \mathcal{S} = \{ (\boldsymbol{v_i}, \boldsymbol{p_i})\}_{i=1}^n$ contains paired videos and corresponding paragraphs that are describing the videos. These are embedded in a bottom-up manner that matches the hierarchical structure of the data. In the video pathway, the frames $f^{ij}_k$ of the clip $ \boldsymbol{c^i_j} = \{ f^{ij}_1, \ldots, f^{ij}_{n_{{ij}}} \} $ are encoded using a pre-trained CNN, followed by a GRU-type RNN $\operatorname{GRU}_c$ which captures the temporal context of the $n_{ij}$ frames of video clip $\boldsymbol{c^i_j}$:
\begin{equation}\label{key}
\begin{aligned}
( \bar{f}^{ij}_1, \ldots, \bar{f}^{ij}_{n_{^{ij}}} ) =\operatorname{GRU}_{c} \left( \operatorname{CNN}(f^{ij}_1), \ldots, \operatorname{CNN}(f^{ij}_{n_{ij}})  \right). % \,.\\
\end{aligned}
\end{equation}
Similarly, the $m_{ij}$ words  $w^{ij}_k$ of the sentence $ \boldsymbol{s^i_j} = \{ w^{ij}_1, \ldots, w^{ij}_{m_{ij}} \} $ are encoded by GloVe~\cite{pennington2014glove} followed by an RNN 
\begin{equation}
(\bar{w}^{ij}_1, \ldots, \bar{w}^{ij}_{m_{^{ij}}} ) =\operatorname{GRU}_s  \left( \operatorname{GloVe}(w^{ij}_1), \ldots, \operatorname{GloVe}(w^{ij}_{m_{{ij}}} )  \right),
\end{equation}

where $\operatorname{GRU}_s$ is the sentence RNN whose sequence is composed of words. Both GRUs have a single layer of GRU cells and an output of size $d_e$.

\begin{figure}[t]
	\includegraphics[width=0.9\linewidth]{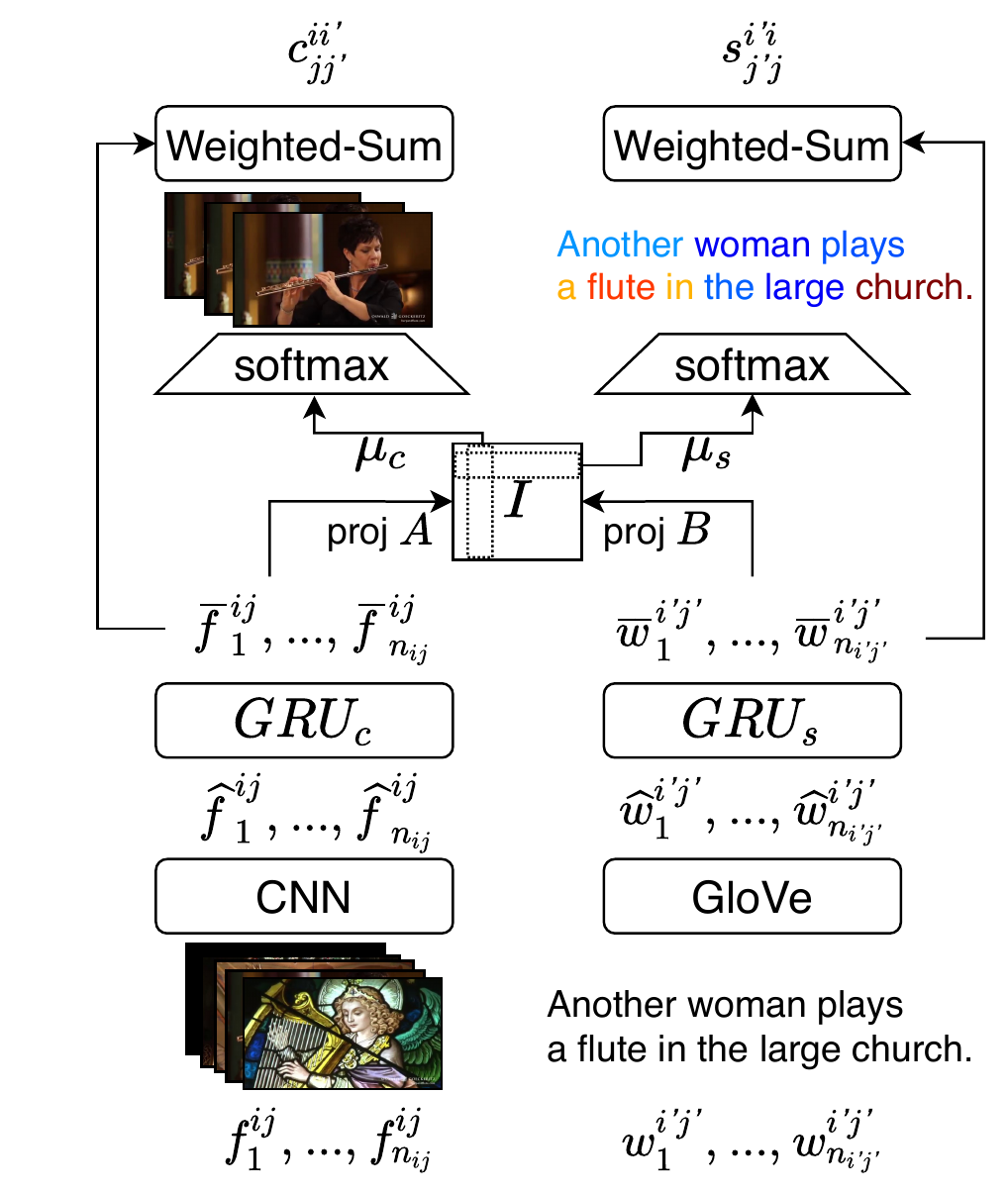}
    \caption{An overview of our first hierarchy for clip and sentence embedding computation. The frames $f^{ij}_k$ of the video $\boldsymbol{v_i}$ and the clip $\boldsymbol{c^i_j} $ and the words $w^{i'j'}_k$ of the paragraph $\boldsymbol{p_{i'}}$ and the sentence $s_{j'}^{i'}$,  are processed in the context of each other. A similarity matrix $I$ is computed based on a dot-product. We calculate interaction potential scores $\mu_c$ and $\mu_s$ by marginalization. We then produce the conditioned embeddings $ c^{ii'}_{jj'} $ and $ s^{i'i}_{j'j}$ based on the interaction potential score.}
    \label{fig:archa}
\end{figure}

Up until this stage, the video and text pathways have been independent. Next, we apply our conditioning mechanism in order to highlight individual frames that are related to the words of the sentence and vice versa as illustrated in \figref{fig:archa}. For this purpose, we define the following matrix of interaction scores, which is given as a normalized dot product:
\begin{equation}\label{interaction}
\operatorname{I}_{kk'}^{ii'jj'} =  \left(\frac{A \bar{f}^{ij}_k}{\norm{A  \bar{f}^{ij}_k}}\right)^\top \left(\frac{B \bar{w}^{i'j'}_{k'}}{\norm{B \bar{w}^{i'j'}_{k'}}}\right), 
\end{equation}
where $\operatorname{I} \in \mathbb{R}^{n_{ij}\times m_{i'j'}}$  and $A \in \mathbb{R}^{d_e \times d_e},  B\in \mathbb{R}^{ d_e \times d_e} $ are learned matrices that re-embed the frames and the words to their joint space $\cR^{d_e}$. The indices of the video/clip/frame $i,j,k$ are distinguished from those of the paragraph/sentence/word $i',j',k'$.

We define the marginal interaction potential over frames and words to be 

$\rho_{c}(k | \boldsymbol{c^i_j}, s^{i'}_{j'}) = \sum_{k'} \operatorname{I}_{kk'}^{ii'jj'}$,  $\rho_{s}(k' | \boldsymbol{c^i_j}, s^{i'}_{j'}) = \sum_{k} \operatorname{I}_{kk'}^{ii'jj'}$.

The potentials are then transformed, using softmax, to a conditional probability distribution over frames and words given their clip and sentence. 
\begin{eqnarray}\label{prob}
\begin{aligned}
\mu_{c}(k| \boldsymbol{c^i_j}, s^{i'}_{j'}) \propto\exp(\rho_{c}(k | \boldsymbol{c^i_j}, s^{i'}_{j'})), \\ \quad \mu_{s}(k'| \boldsymbol{c^i_j}, s^{i'}_{j'}) \propto\exp(\rho_{s}(k' | \boldsymbol{c^i_j}, s^{i'}_{j'})).
\end{aligned}
\end{eqnarray}

With these marginal probabilities, we entangle the clips and the sentences by reducing their representation, namely, summarize the clip representation by combining its sentence-related frames and the sentence representation by its clip-related words:  
\begin{eqnarray}\label{poten}
\begin{aligned}
c^{ii'}_{jj'} = \sum_{k=1}^{m_{{ij}}} \mu_{c}(k | \boldsymbol{c^i_j}, s^{i'}_{j'}) \bar{f}^{ij}_k, \\ \quad
s^{i'i}_{j'j} = \sum_{k'=1}^{n_{{i'j'}}} \mu_{s}(k' | \boldsymbol{c^i_j}, s^{i'}_{j'}) \bar{w}^{i'j'}_{k'} .
\end{aligned}
\end{eqnarray}
Since these representations are derived from the joint interactions space of clips and sentences, the resulting embeddings of $c^{ii'}_{jj'}$ and $s^{ii'}_{jj'} $  are conditioned on each other. The next section describes an additional encoding hierarchy that enables video and paragraph matching.

\subsection{Joint Video and Paragraph Encoding}
\label{sec:video_para}
We find it easy to extend out conditioned representation to the matching of a video and a paragraph. Given interaction-conditioned representation. \Ie, video's clips $\{ c^{ii'}_{^{11}}, \ldots,  c^{ii'}_{^{{n_{ii'}n_{i'i}}}} \}$ and paragraph's sentences  $\{s^{i'i}_{11}, \ldots, s^{i'i}_{^{{n_{i'i} {n_{ii'}}}}} \} $. We encode them with a pair of GRUs where $n_{ii'} = \max(n_{i'}, n_{i})$
\begin{equation}\label{eq:gru}
\begin{aligned}
(\hat c^{ii'}_1, \ldots, \hat c^{ii'}_{n_{ii'}} ) = \operatorname{GRU}_v\left( (  c^{ii'}_{^{11}}, \ldots,  c^{ii'}_{^{{n_{ii'}} }}) \right), \\ \quad (\hat{s}^{i'i}_{1}, \ldots, \hat{s}^{i'i}_{n_{ii'}} ) = \operatorname{GRU}_p\left(  (s^{i'i}_{11}, \ldots, s^{i'i}_{^{n_{i'i}n_{ii'}}} )\right).
\end{aligned}
\end{equation}
In case the number of clips and the number of sentences are not equal, we pad with a zero vector the missing representations. 

Notably, to this point, each clip and sentence interact locally. In other words, objects in a clip only interact with words in the corresponding sentence and vice versa.  However, we find it beneficial to also condition the video and paragraph in a global context, \ie, words/objects interact with non-corresponding sentence/clip. To this end, we sample $n_f$ frames from the entire video, along with the words of the entire paragraph. We then apply the same process as in Eq.~\ref{interaction} - Eq.~\ref{poten}, treating the sampled frames and words as a single clip and a single sentence. The only difference is that we employ a different set of matrices,  instead of $A$ and $B$ (Eq.~\ref{interaction}), which we denote by $A_0 \in \mathbb{R}^{d_e \times d_e}$ and $B_0\in \mathbb{R}^{ d_e \times d_e}$. We use the interaction-conditioned global video embeddings, denoted as $v^0_{ii'}$ as the RNN $\operatorname{GRU}_v$'s initial state.  Similarly, we denote the global paragraph embedding $p^0_{ii'}$, which is used to initialize the RNN $\operatorname{GRU}_p$.

Finally, to co-dependently embed the entire video $\boldsymbol{v_i}$ and the entire paragraph $\boldsymbol{p_{i}}$ , max-pooling is applied across all video clips and paragraph sentences, respectively: 
\begin{equation}\label{}
\begin{aligned} v_{ii'} =\operatorname{MaxPool}\left( \hat{c}^{ii'}_1, \ldots, \hat{c}^{ii'}_{n_{ii'}} \right) ,\\ \quad   p_{i'i}=\operatorname{MaxPool}\left( \hat{s}^{i'i}_1, \ldots, \hat{s}^{i'i}_{n_{ii'}} \right).
\end{aligned}
\end{equation}
The architecture of the processing at the video and paragraph level is summarized in \figref{fig:archb}.

\subsection{Training} The loss employed is based on the triplet loss of the video and paragraph embeddings together with an auxiliary loss that also takes the form of a triplet loss. Recall that for matching videos and paragraphs $(\boldsymbol{v_i},\boldsymbol{p_i})$, the $j$-th clip $\boldsymbol{c^i_j}$ matches the $j$-th sentence $\boldsymbol{s^i_j}$. For this purpose, we learn another pair of matrices $U$ and $V$ to define a match score:

\begin{figure}[t]
\label{fig:archb}
\includegraphics[width=.9\linewidth]{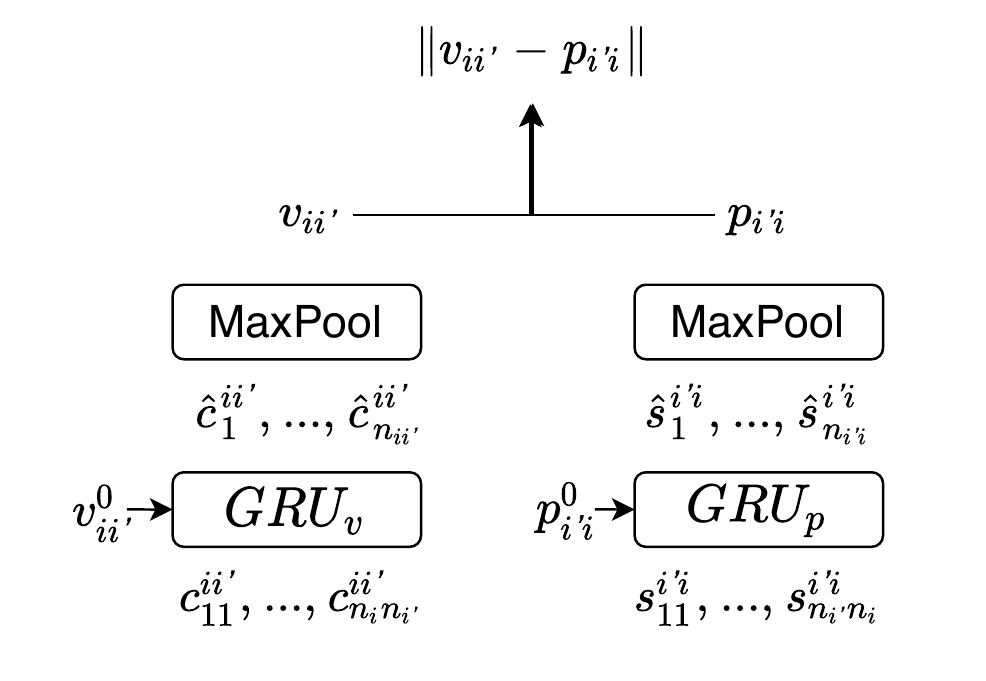}
\caption{Our second hierarchy for the embedding and then matching a video and a paragraph based on the interaction-conditioned representations of the video's clips and the paragraph's sentences. Note that we employ this hierarchy only in the case of paragraph and video matching.}
\end{figure}

\begin{table}[t!]
	\begin{center}
	\resizebox{\linewidth}{!}{%
	\begin{tabular}{llcccccc} 
	\toprule
	&& \multicolumn{3}{c}{Text2Vid} & \multicolumn{3}{c}{Vid2Text}  \\ 	
				\cmidrule(lr){3-5} 	\cmidrule(lr){6-8}    

		Dataset&Method&R@1&R@5&MdR&R@1&R@5&MdR\\
		\midrule
         &\textcolor{gray}{MMT$\ddagger$ }~\cite{gabeur2020multi}&28.7\%&61.4\%&3.3&28.9\%&61.1\%&4.0 \\
		\cmidrule(lr){2-8} 
		\multirow{2}{*}{ActivityNet}&FSE~\cite{zhang2018cross}&18.2\%&44.8\%&7.0&16.7\%&43.1\%&7.0  \\
        &HSE~\cite{zhang2018cross}&20.5\%&49.3\%& - & 18.7\% & 48.1\% & - \\
         &CE$\ddagger$ ~\cite{liu2019use}&18.2\%&47.7\%& 6.0& 17.7\%&46.6\% & 6.0\\
         &MMT$\ddagger$ ~\cite{gabeur2020multi}&22.7\%&54.2\%&5.0&22.9\%&54.8\%&4.3 \\
         &clipBERT$\dag^{*}$  \cite{lei2021less} &  21.3\% & 49.0\% & -  & - & - & -
		\\
		\cmidrule(lr){2-8} 
		&Ours & \textbf{25.4\%} & \textbf{59.1\% } & \textbf{3.0}& \textbf{26.1\%} & \textbf{60.0\%}  & \textbf{3.0}\\
        \midrule

		\multirow{4}{*}{LSMDC}
		&\textcolor{gray}{MMT$\ddagger$} \cite{gabeur2020multi}& 12.9\%  &  29.9\%  & 19.3  & 12.3\%   &  28.6\%  & 20.0  \\
				\cmidrule(lr){2-8} 
		&HT100 ~\cite{miech2019howto100m}  & 7.1\%  & 19.6\%   & 40.0 &  - &  - & - \\
		&JFusion \cite{yu2018joint}& 9.1\% & 21.2\% & 36.0  & - &  - &  - \\
		&HSE~\cite{zhang2018cross}  & 9.5\%  & 25.8\%  & 23.0 &  9.7\% &  25.1\%  & 23.0 \\
		&MMT$\ddagger$ \cite{gabeur2020multi}& 13.2\%  &  29.2\%  & 21.0  & 12.1\%   &  29.3\%  & 22.5  \\
		\cmidrule(lr){2-8} 
		&Ours & \textbf{14.9\%} &  \textbf{33.2\%} & \textbf{19.0} & \textbf{15.3\%}  &  \textbf{34.1\%} & \textbf{18.0} \\
		\midrule 
		\multirow{5}{*}{MSR-VTT}
		&\textcolor{gray}{HT}~\cite{miech2019howto100m} & 14.9\% & 40.2\% & 9.0 & - & - & - \\
		&\textcolor{gray}{MMT$\ddagger$} \cite{gabeur2020multi}&26.6\%  &  57.1\% & 4.0   &27.0\% &  57.5\%   &  3.7 \\
		&\textcolor{gray}{ActBERT}~\cite{zhu2020actbert} & 16.3\% &  42.8\% & 10.0 & - & - & - \\
		&\textcolor{gray}{HERO}~\cite{li2020hero}&16.8\% &  43.4\% & -& - & - & - \\
		&\textcolor{gray}{HERO}$\dag$~\cite{li2020hero}&20.5\% & 47.6\% & -& - & - & - \\
        \cmidrule(lr){2-8} 
        &Dual Encoding \cite{dong2019dual}& 7.7\% & 22.0\% &  32.0 & 13.0\% &  30.8\%  &  15.0 \\
		&HGR \cite{chen2020fine}& 9.2\% &  26.2\% &   24.0 & 15.0\% &  36.7\% & 11.0  \\
		&JSFusion \cite{yu2018joint}& 10.0\% & 31.2\%&  13.0 & - &  - &  - \\
		&Cues \cite{mithun2018learning}& 12.5\% &  32.1\% &  16.0 & 7.0\% &  20.9\%   & 38.0  \\

		&HSE~\cite{zhang2018cross} 	& 17.9\%  &  45.0\% & 23.0 & 18.5\% & 45.2\%   & 23.0  \\
		&MMT$\ddagger$ \cite{gabeur2020multi}&24.6\%  &  54.0\% & 4.0   &21.1\% &  49.4\%   &  6.0 \\
		&clipBERT$\dag^*$ \cite{lei2021less} &  22.0\% &  46.8\%  & -  & - & -   & -
		\\
		\cmidrule(lr){2-8} 
		&Ours&  \textbf{26.0\%} & \textbf{ 56.7\%}  & \textbf{3.0}  & \textbf{26.7\%}  & \textbf{56.5\%}   & \textbf{3.0}   \\
        \bottomrule 
    \end{tabular}}\vspace{-5pt}
	\end{center}
	 \caption{Video/Sentence retrieval on ActivityNet~\cite{krishna2017dense}, LSMDC~\cite{rohrbach2015dataset}, and MSR-VTT~\cite{xu2016msr}. Shown is the recall at a certain number of retrievals and MdR=median rank. We gray out models that used large-scale pre-training on HowTo100M~\cite{miech2019howto100m}.  We mark with $\ddagger$ models that used  appearance, scene, motion, face, audio, OCR, ASR features from 11 different models. $\dag$ indicates models that used appearance and motion. (*) denotes models use 2-second clips instead of the default 1-second clips.}\vspace{-13pt} \label{tab:sentence_ret}
\end{table}
\begin{equation}\label{csinteraction}
\operatorname{M}(x, y)%(\bar{f}^{ij}_k, \bar{w}^{i'j'}_{k'}) 
=  \left(\frac{U x}{\norm{U  x}}\right)^\top \left(\frac{V y}{\norm{V y}}\right).
\end{equation}
We note that for the video/paragraph case, $U$ and $V$ are only used during training and do not play a role in matching in inference time. In total, we learn three normalized dot-product cross-domain interactions. In Eq.~\ref{interaction} we learn $A$ and $B$ at the frame and word level. Similarly, we learn $A_0$ and $B_0$ for matching frames sampled from entire-video and words sampled from the entire-paragraph. Last, we learn $U$ and $V$ at the clip and sentence level. 

In this metric learning task, positive pairs are obtained by co-embedding corresponding  video and paragraphs (index $i$) using the corresponding clip/sentence (index $j$). Negative pairs are obtained using either unmatching clips and sentences, introducing a second index $j'\neq j$, or using entirely different video and paragraph, denoted by the index $i'\neq i$. 
The mini-batches are constructed by considering $b$ pairs of matching videos and paragraphs. Then, all non-matching pairs in the mini-batch are considered as negative samples. The loss is given by:
\begin{multline}
    \hspace{-0.5cm}
    {\mathcal{L}}=\sum_{i=1}^b \sum_{i'\neq i} \text{ReLU}(\| v_{ii} - p_{ii} \|-\|v_{ii'} - p_{i'i} \| + \alpha_1) \\
       +
       \sum_{i=1}^b  \sum_{j=1}^{n_{i}}\sum_{j'\ne j} \text{ReLU}(\operatorname{M}(c^{ii}_{jj}, s^{ii}_{jj})-\operatorname{M}(c^{ii}_{jj'}, s^{ii}_{j'j})+\alpha_2)\\ 
      + \sum_{i=1}^b \sum_{i'\neq i} \sum_{j=1}^{n_{i}}\sum_{j'=1}^{n_{i'}} \text{ReLU}(\operatorname{M}(c^{ii}_{jj}, s^{ii}_{jj})-\operatorname{M}(c^{ii'}_{jj'}, s^{ii'}_{j'j})+\alpha_2),
\end{multline}
where $\alpha_1$ and $\alpha_2$ are two margin parameters. The loss has three parts. The first compares the representation of entire video $i$ and considers the matching paragraph, and an unmatched paragraph obtained as the matching paragraph of another video $i'$. Note that this term only applies in the case of video/paragraph matching. The second term is at the level of clips and sentences. It considers matching clips and sentences (matching video and paragraph, both with index $i$, and same clip and sentence, both with sentence $j$) and unmatching sentences of the same paragraph (same $i$ different $j'\neq j$). The third term is similar, only the unmatched sentences are taken from an unmatched paragraph (different $i'\neq i$ and all possible $j'$). Note, we pad with zeros the long video/paragraph to have an equal length as the shorter video/paragraph.

\subsection{Test Time Similarity Computation} To evaluate our video retrieval capabilities, we compute a similarity score $ \operatorname{S} $ between every clip/sentence or video/paragraph pair. Given a clip $ \boldsymbol{c_i} $,  we use the trained matching operator $M$ (see \eqref{csinteraction}), \ie,
\begin{equation}
\operatorname{S}(\boldsymbol{c^{i}_{j}}, \boldsymbol{s^{i'}_{j'}}) =\operatorname{M}\left(c^{ii'}_{jj'}, s^{i'i}_{j'j}\right).
\end{equation}
Note, for video/sentence matching, all clips are concatenated into one clip. 
In the case of video paragraph retrieval, given a paragraph $\boldsymbol{p_{i'}} $, we calculate the similarity score based on a dot product between their dependent representation as follows:
\begin{equation}
    \operatorname{S}(\boldsymbol{v_i}, \boldsymbol{p_{i'}}) = \exp(-\|v_{ii'} - p_{i'i}\| ).
\end{equation}

\section{Experiments}
We evaluate our video and text matching method on a variety of datasets containing video clips with corresponding descriptions. To demonstrate the versatility of our model, we perform a comprehensive set of experiments in a diverse set of domains, such as movies, personal collections, and commercials. In addition to the matching task, we also explore weakly supervised localization in video and even video-guided machine translation.
\begin{table}[t!]

	{
		 %\resizebox{\columnwidth}{!}{%
		 	\begin{center}
	\resizebox{\linewidth}{!}{%
	\begin{tabular}{lccccccc} 
	\toprule
	&& \multicolumn{3}{c}{Text2Vid} & \multicolumn{3}{c}{Vid2Text}  \\ 	
				\cmidrule(lr){3-5} 	\cmidrule(lr){6-8}    
		Dataset&Method& R@1 & R@5 &MdR& R@1 & R@5  &MdR  \\
		\midrule
		\multirow{4}{*}{ActivityNet}&LSTM-YT~\cite{venugopalan2015sequence}  & 0.0\% & 4.0\%  & 102.0 &0.0\% & 7.0\%  & 98.0       \\
		&No Context~\cite{venugopalan2014translating} & 5.0\% &  14.0\%  & 78.0 & 7.0\% & 18.0\% & 56.0  \\
		&Dense full~\cite{krishna2017dense}&14.0\%&32.0\%&34.0& 18.0\%&36.0\% & 32.0 \\
		&FSE~\cite{zhang2018cross} & 18.2\% &  44.8\% &7.0& 16.7\% & 43.1\%& 7.3  \\
		&HSE~\cite{zhang2018cross} & 44.4\% &  76.7\%   &2.0& 44.2\% & 76.7\%& 2.0  \\
		\cmidrule(lr){2-8} 
		&Ours  & \textbf{58.8\%} &  \textbf{89.2\%}  &\textbf{1.0} & \textbf{58.2\%} & \textbf{88.1\%}  & \textbf{1.0} \\
	    \midrule
		\multirow{3}{*}{DiDeMo}&S2VT~\cite{venugopalan2014translating}   & 11.9\% &  33.6\% &13.0 & 13.2\%  & 33.6\% & 15.0  \\
		&FSE~\cite{zhang2018cross}  & 13.9\% & 36.0\%. & 11.0 & 13.1\% & 33.9\% & 12.0 \\
		&HSE~\cite{zhang2018cross}  & 30.2\% & 60.5\% & 3.3 & 30.1\% & 59.2\%  & 3.0 \\
		\cmidrule(lr){2-8} 
		&Ours & \textbf{38.5\%}  & \textbf{72.7\%}   &\textbf{2.7} & \textbf{38.7\%}  & \textbf{100.0\%}  & \textbf{2.4}   \\
		\midrule
		\multirow{2}{*}{YouCook2}&HSE~\cite{zhang2018cross}  & 43.5\% &  80.7\% &2.0 & 43.8\% & 83.4\% & 2.0\\
		\cmidrule(lr){2-8} 
		&Ours & \textbf{49.1\%}  & \textbf{87.2\%}  &\textbf{1.0} & \textbf{51.3\%} & \textbf{90.0\%}    & \textbf{1.0}    \\
		\midrule
		\multirow{2}{*}{MSR-VTT}&HSE~\cite{zhang2018cross}  & 32.9\%& 64.0\% &  3.0 & 32.5\% & 63.7\%  & 3.0 \\
		\cmidrule(lr){2-8} 
		&Ours & \textbf{38.0\%}  & \textbf{72.5\%}   &\textbf{2.0}& \textbf{37.8\%} & \textbf{72.0\%}   & \textbf{2.0}     \\
			\bottomrule
	\end{tabular}}\vspace{-5pt}
	\end{center}
	\caption{Video/Paragraph retrieval on ActivityNet~\cite{krishna2017dense}, DiDeMo~\cite{anne2017localizing} , YouCook2~\cite{ZhXuCoCVPR18} and MSR-VTT~\cite{xu2016msr}. Shown is the recall at a certain number of retrievals and MdR=median rank.}\vspace{-13pt}\label{tab:baselineresults}}%}
\end{table}
\noindent\textbf{Datasets:} We employ five different datasets, each having a different distribution of videos and a different descriptive style. For example, ActivityNet focuses on actions (``The girl dances around the room while the camera captures her movements.'') while YouCook descriptions capture the order of very specific actions (``combine lemon juice sumac garlic salt and oil in a bowl''). The datasets we employ are listed next. {\bf ActivityNet Dense Caption~\cite{krishna2017dense}} is a human activities video dataset, where each video has multiple events. The events are separated into clips, and each clip has a corresponding description. There are 849 video hours with 100,000 total sentences, giving rise to 10,009 videos with matching paragraphs for the training set, and 4,917/4,885(val1/val2) for the validation set. We follow previous works and report our results on val1. {Results on val2 are available in the supplementary.} {\bf Distinct Describable Moments (DiDeMo)~\cite{anne2017localizing}} is collected from diverse personal videos and  include everyday indoor scenes, nature scenes, events etc. The dataset consists of over 10,000 unedited videos. Each video comes with several actions described in natural language. The data is split into 8,395 training videos, 1,065 validation videos and 1,004 test videos. We report our results on the test-set.  {\bf YouCook2~\cite{ZhXuCoCVPR18}} contains 2,000 long untrimmed videos capturing the preparation of 89 cooking recipes, where each distinct recipe has 22 videos on average. The videos are provided with temporal boundaries and description sentences in imperative English. The dataset is split into a train set of size 1,500 and a test set with 500 videos. {\bf Vatex~\cite{wang2019vatex}}  contains  41,250 videos with 825,000 multilingual (English and Chinese) captions and includes over 600 fine-grained human activities.
\textbf{MSR-VTT}~\cite{xu2016msr} contains a total of 10K videos, each having 20 text descriptions. This dataset's standard split comprises 6,576 videos for training, 497 videos for validating, and 2,990 videos for testing. \textbf{LSMDC}~\cite{rohrbach2015dataset} contains 118,081 short videos extracted from 202 movies, each described by a text caption extracted from the movie or the audio description. The test set is composed of 1000 videos.

\subsection{Video/Sentence Retrieval}\vspace{-2pt}
In this task, we match a single sentence and a video without clips.  In datasets with clips, we treat each video as a single clip. Similarly, we concatenate all the captioning sentences into a single caption. In \tabref{tab:sentence_ret}, we compare our method the state-of-the-art baselines. The baselines include the method of Liu~\etal~\cite{liu2019use}, who suggested using many video-expert networks, such as speech, audio, motion, OCR, appearance, and face detection. Another recent baseline is the Multimodal Transformer (MMT)~\cite{gabeur2020multi}, which encodes all the expert networks' representations jointly with self-attention. Our method, which considers only the appearance and text embeddings and does not use external knowledge, is much more straightforward, and also outperforms all those methods that did not use conditioned embeddings.

The most recent work clipBERT~\cite{lei2021less} study different frame sampling strategies.  The approach we propose outperforms clipBERT on both ActivityNet and MSR-VTT by 4\%, despite clipBERT's use of motion features. Notably, our technique outperforms pre-trained versions except for MMT, which also relies on many video-experts networks.
 
We also compare the efficiency of our architecture in terms of number of parameters. In total, our model has only 40.2mil parameters, while clipBERT, a Transformer-based network, has 199.3mil parameters. Additionally, MMT that considers various experts networks has 133.3mil parameters.\vspace{-5pt}

\subsection{Video/Paragraph Retrieval}\vspace{-2pt}
Our next task is to match videos with descriptive paragraphs. Compared to the video/sentence retrieval, this variant uses clips segmentation and corresponding sentences. Each corresponding clip and  sentence are conditioned separately. Note that we also employ global conditioning (\ie, the global video and paragraph embeddings). The second hierarchy is used to handle the conditioned clips and sentences representations (see \secref{sec:video_para}).  \tabref{tab:baselineresults} reports the results of our method in comparison to the results collected from the literature as well as HSE results we ran, ourselves, on YouCook2 , and MSRT-VTT.  HSE~\cite{zhang2018cross} also uses hierarchy encoding. Our loss is  considerably simpler than that of HSE. Moreover, our approach leverage conditioned embeddings. Our method obtains an improvement in R@1 over HSE result in both matching directions of 14\% in ActivityNet, 9.6\% in DiDeMo, 7.5\% in YouCook2 and 5.3\% in MSR-VTT.\vspace{-10pt}

\begin{table}
\resizebox{\columnwidth}{!}{
\begin{tabular}{lcccc}%{@{}l@{~}c@{~~}cc@{~~}c@{}} 
	\toprule
	& \multicolumn{2}{c}{Text2Vid} & \multicolumn{2}{c}{Vid2Text}\\
	\cmidrule(lr){2-3} 	\cmidrule(lr){4-5}  
	{Model}&  R@1 & R@5 &  R@1 & R@5 \\
	
	\midrule
	%& \multicolumn{9}{c}{Video to Paragraph} \\
	%HSE & 44.2\% & 76.6\% & 97.0\% \\
		\hline\multicolumn{5}{c}{\textbf{architecture}}    \\\hline 
	w/o 2nd H&  33.0\% & 66.7\% & 32.2\%  & 65.8\% \\ 
	w/o M match & 57.2\%  & 87.9\% & 56.9\%  & 87.1\% \\   
	w/o global & 57.2\% & 87.9\% & 57.0\%  & 87.4\% \\ 
	\hline\multicolumn{5}{c}{\textbf{conditioning}}    \\\hline 
    w/o attn & 49.1\% & 80.6\%& 47.4\%& 80.2\%\\
    w/o frames attn &  52.7\% & 82.0\% & 51.3\% & 82.3\% \\ % 98.7\%  
	w/o sentence attn & 54.6\%  &  84.5\%  & 52.9\% & 83.4\%  \\ 
	%Shared High-Hierarchy weights &  54.3\% & 85.1\% & 100.0\%  &  54.7\% & 86.0\% & 100.0\%    \\
	w/o global attn &  56.6\% & 87.3\% &  56.2\% & 86.9\% \\ 
	alternative attn   & 56.1\% & 87.3\% & 55.6\%  & 86.8\% \\ %& 100.0\%   
	\hline\multicolumn{5}{c}{\textbf{loss}}    \\\hline 
    w/o clip match & 52.9\% &  84.1\% & 52.3\%  & 82.2\% \\

%Dim = 256	51.80%	82.10%		50.50%	81.70%	
%Dim = 512	55.10%	86.20%		54.60%	85.70%	
%Dim = 1024	54.70%	86.70%		54.00%	86.00%	
	\hline\multicolumn{5}{c}{\textbf{embedding-dimension}}    \\\hline 
    $d_h=256$ & 54.9\% & 87.1\%&	 53.9\%	&85.7\% \\
    $d_h=1024$ & \textbf{58.9\%}	& \textbf{90.1\% } & 57.8\%	& 87.8\%\\
    \cmidrule(lr){1-5} 
	Full &  58.8\% & 89.2\% &  \textbf{58.2\%} & \textbf{88.1\%}  \\ % \textbf{100.0\%}   
	\bottomrule
\end{tabular}}
\caption{An ablations study on ActivityNet.}\vspace{-17pt}\label{tab:ablation}
\end{table}

\begin{figure*}[t]
	\centering
	\includegraphics[width=1\linewidth]{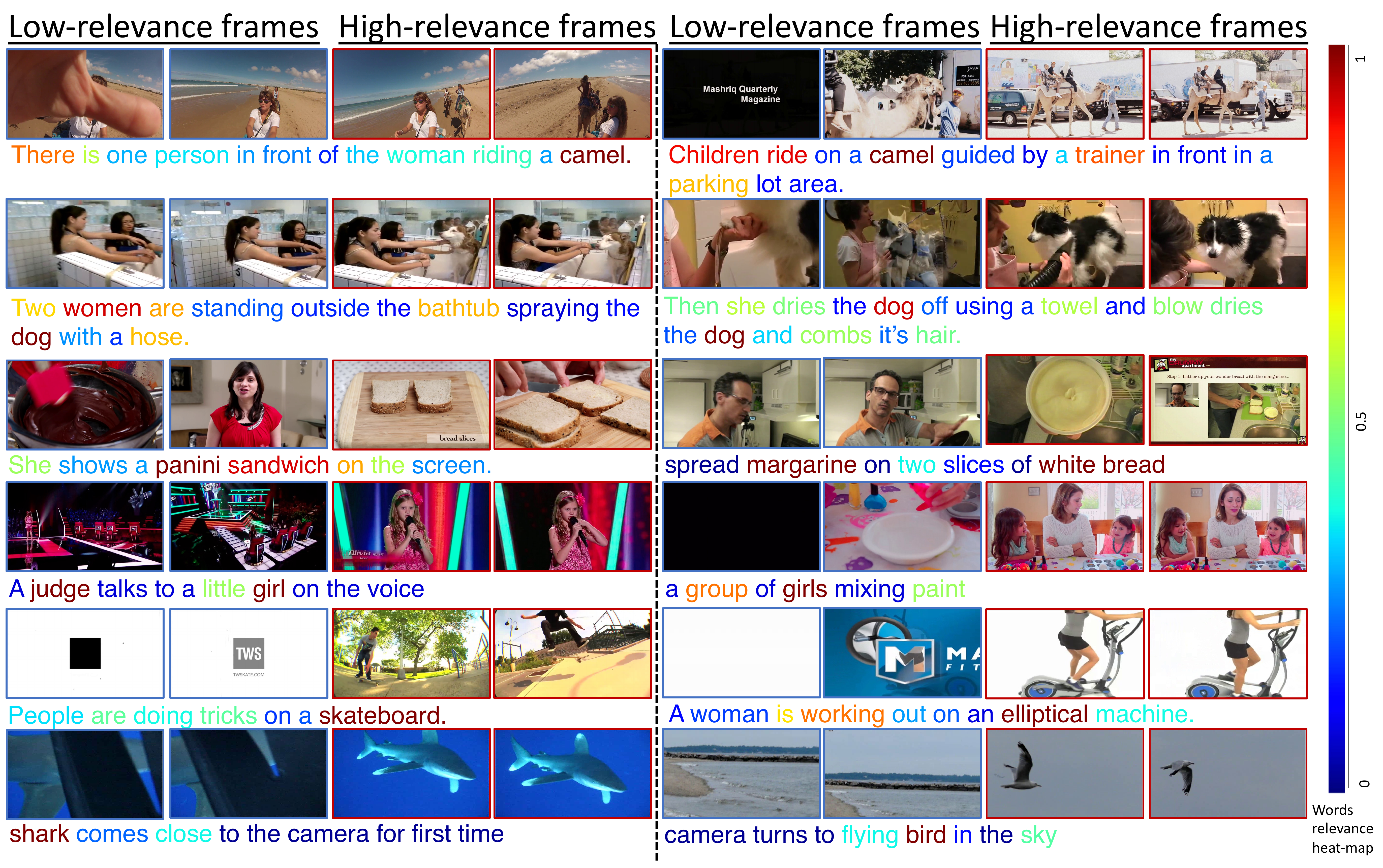}\vspace{-5pt}
	\caption{Illustration of queries, relevancy of frames $\mu_c$  and words $\mu_s$ derived from \equref{prob}.  We show results on ActivityNet (1st, 2nd, 5th rows), YouCook2 (3rd row), DiDeMo (6th row), MSR-VTT (4th row). Each row displays two queries and a clip from the retrieved video. In the clip, we show two of the most relevant frames to the query (marked with red borders) and two of the least relevant frames (marked with blue borders). We also illustrated relevant words to the clip using color maps. Note, our approach also serves the additional benefit of describing the specific frames the model uses for retrieval.} \vspace{-18pt}%Note, the attention scores are normalized potentials.}
	\label{fig:attention}
\end{figure*}

\subsubsection{Ablation study} \vspace{-5pt}
In \tabref{tab:ablation} we assess several design choices evaluated on video and paragraph retrieval setup: 1) We examine the importance of different architecture components.    In `w/o 2nd H', we used only one hierarchy by treating the output of our global embedding as the conditioned sentence and video representations (\ie, $v^0_{ii'}$, $p^0_{ii'}$). While the performance dropped significantly, this score is impressive because the model does not take into account the events' temporal locations (\ie, clips). In `w/o M match', we evaluate our model with a normalized dot-product instead of our trainable matching module $M$. Last, in `w/o-global', we do not use a global video and paragraph embeddings to initialize $GRU_v$ and $GRU_p$. In this case, the performance drops by 1\% on R@1.   Overall, we show that all the integral components in our architecture contribute to the performance. 2) We assess our conditioning approach. We treat it as a form of co-attention, \ie, picking the relevant frames and words. In `w/o-atten', we show the importance of attention, omitting the attention leads to a significant drop in performance (47.4\% \vs 58.2\%). In `w/o text attn', we infer the clip's attention as usual, but instead of text attention, we performed a max-pooling across the sentence representation. As a result, the performance drops by 5.3\% on R@1. Similarly, in `w/o frame attn', we removed the clips' attention, which leads to a significant performance drop of 7\% on R@1. In `w/o global attn', we use max polling for the global vector instead of attention, which leads to a drop of 2\% on R@1. In `alternative attn', we replace our attention with that of~\cite{yuan2019find}, which leads to a 2.6\% drop on R@1.  Additionally, our version runs 36\% faster. 3) Removing the auxiliary clip and sentence matching drops the performance by 7.2\% on R@1. 4) We change our main embedding parameter $d_h$. It is evident that our method is stable to this parameter in terms of Recall@k metric and it outperforms state of the art (Tab.~\ref{tab:baselineresults}) for multiple values.\vspace{-15pt}

\subsubsection{Qualitative analysis} \vspace{-5pt}

In \figref{fig:attention} we use the interaction scores (see \equref{prob}) to identify the relevant frames and words selected by conditioning. Our approach simplifies explanations since finding the frames responsible for retrieving a specific long video can be challenging.
In the first step, we highlight the words' relevance through a heatmap. We find that most of the relevant words are grounded nouns. For example, in the first row, the relevant words in the query are `children,' `camel,' `trainer,' `parking.' These are useful words for visual matching. 
Next, we randomly selected a clip from the retrieved video and displayed the top two frames that most closely match the query, as indicated by the red border. The bottom two frames, are marked by blue borders. The bottom frames are usually irrelevant to the query.  For instance, in the 2nd row, on the left side, the query is ``... spraying the dog,'' but the bottom frames do not depict any dog. We also find the frames to be uninformative. In the first row, on the left, the woman's hand hides most of the details. In addition, clips often have frames with production logos, \eg, the bottom frames in 5th-row videos.%, when the query is ``... tricks on a skateboard'',  the top frames are skateboard tricks .
We also show a complicated case of two similar queries. Notably, both queries in the first row deal with camels; in the second row, both queries deal with dogs; in the third row, both videos deal with bread sandwiches. Despite being similar, our model identified subtle details and retrieved the correct video. The 4th row shows samples from the MSR-VTT dataset. The top 2 frames in the left examples are related to `girl' as highlighted in the query attention, and for the right example, we can see that the top 2 frames contain `group of girls.' In the 3rd row, we show two queries from Youcook2 and find our approach to pick the relevant frames showing the query's ingredients (\eg, a sandwich, margarine, white bread). Lastly, in the 6th row, we show queries from DiDeMo. The attention behaves well by selecting relevant frames, \eg a shark, and a bird.\vspace{-3pt}

\begin{figure}[t]
	\includegraphics[width=1\linewidth]{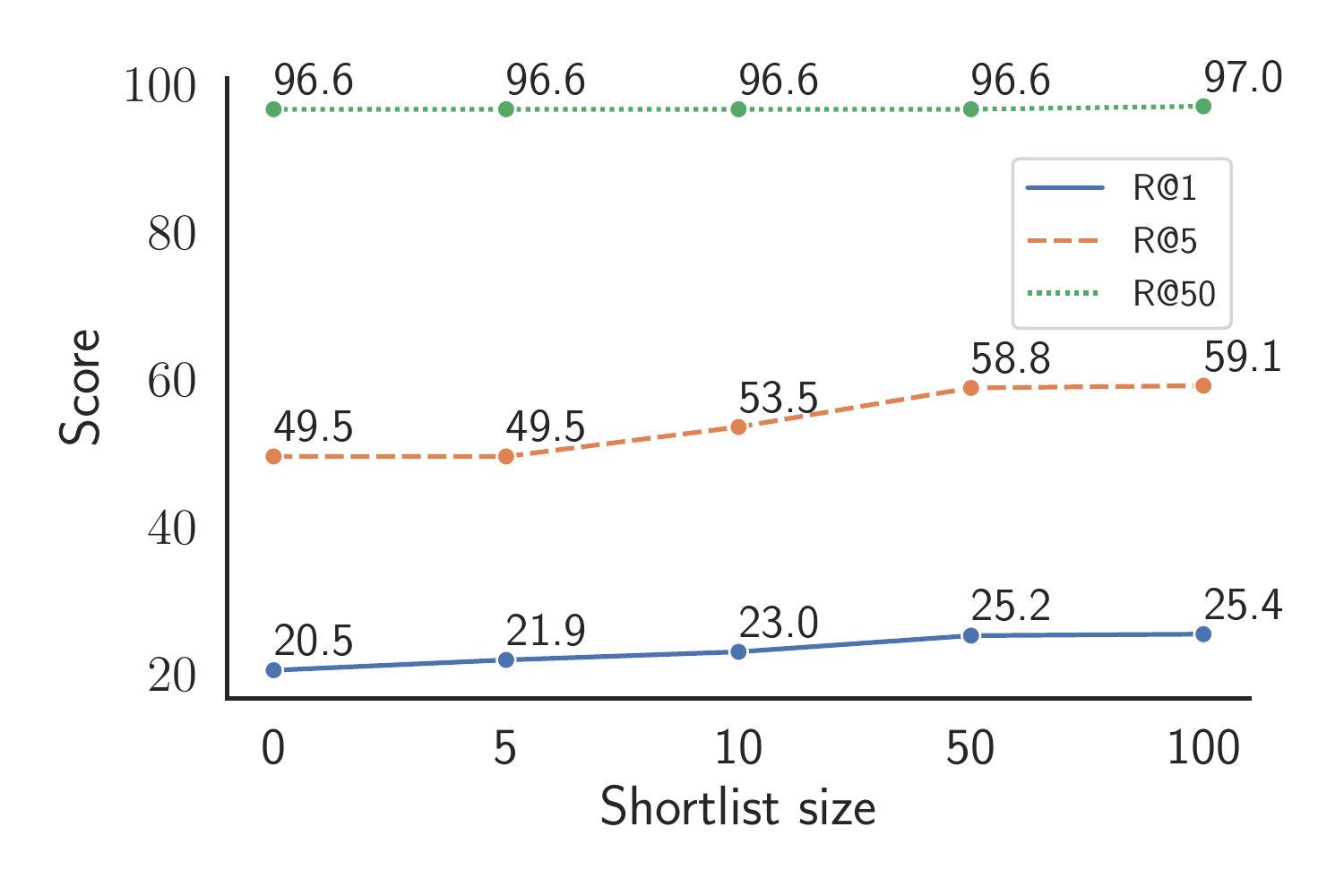}\vspace{-10pt}
	%\captionsetup{justification=centering}
	\caption{Comparison of different shortlist sizes. The purpose of the shortlist set is to allow the user to reduce inference latency. Hence, we use a static retrieval strategy to construct a candidate set. Notably, the performance improves with increasing set size, but the inference latency increases (see \secref{sec:shortlist}). We find the size of 50, which we also used in our experiments, to yield a good trade-off between latency and performance.}\vspace{-15pt}
	\label{fig:shortlist}
\end{figure}

\subsection{Inference time and shortlist}\vspace{-3pt}

\label{sec:shortlist}

For each inference call, our approach requires recomputing the representations. This computational burden can be alleviated through the use of a static method (\eg, HSE~\cite{zhang2018cross}) to retrieve a shortlist of relevant candidates. In \figref{fig:shortlist}, we examine different shortlist sizes and find that a shortlist of 50 already allows most of the performance gain. We measured the inference time on ActivityNet in the paragraph/video task. The inference latency for a single inference call is $\sim$0.4s for a shortlist of 5, $\sim$1.1s for a shortlist of 10, $\sim$1.8s for a shortlist of 50, and $\sim$3.2s for a shortlist of 100. The overall inference time of ActivityNet test set for a shortlist of 50 is $\sim$7 minutes, in comparison to $\sim$50 minutes without a shortlist and $\sim$50s using the HSE method as static approach. Note, inference of advanced static approaches also takes a longer time compared to HSE. For example, MSE~\cite{gabeur2020multi} inference takes about four minutes over the test set.\vspace{-3pt}

\subsection{Weakly Supervised Retrieval}\vspace{-3pt}

In many scenarios, the supervision of the events' temporal location is not available. The weakly supervised setup is similar to the Video/Paragraph retrieval task, excluding the events segmentation. %The weakly supervised retrieval task was proposed to address this issue. This setup 
In \tabref{tab:proposals}, following~\cite{zhang2018cross}, we report results obtained when breaking the paragraph into sentences, and replacing the ground-truth events in the ActivityNet dataset with equal-length non-overlapping temporal segments. For the textual input, in case there are fewer segments than sentences, we truncate the paragraph. If there are more segments than sentences, we repeat the last sentence. The absence of ground-truth events and truncating the paragraph harms the performance (24\% drop on R@1). Still, our model outperforms the previous baseline by 5.0-6.5\% on R@1 depending on the number of segments. \vspace{-2pt}

\begin{table}[t]
%	\parbox{.48\linewidth}{
	\begin{center}%\captionsetup{font=footnotesize}
		%\scalebox{0.72}{}
		\begin{tabular}{llcccc}%{@{}llc@{~~}cc@{~~}c@{}}
			\toprule
			& & \multicolumn{2}{c}{Text2Vid}& \multicolumn{2}{c}{Vid2Text}\\
			\cmidrule(lr){3-4} 	\cmidrule(lr){5-6}  
			N&Alg.& R@1 & R@5& R@1 & R@5 \\
			\midrule
			\multirow{2}{*}{1}&HSE  & 18.0\% & 45.5\% & 16.5\% & 44.9\% \\
			&Ours  & \textbf{21.9\%} & \textbf{49.3\%} & \textbf{21.5\%} & \textbf{49.1\%}\\
			\midrule
			\multirow{2}{*}{2 }&HSE & 20.0\% & 48.9\% & 18.4\% & 47.6\% \\
			&Ours &  \textbf{24.8\%} & \textbf{55.6\%} & \textbf{24.3\%} & \textbf{54.6\%}\\
			\midrule
			\multirow{2}{*}{3}&HSE & 20.0\% & 48.6\%  & 18.2\% & 47.9\% \\
			&Ours  & \textbf{24.9\%} & \textbf{55.8\%} & \textbf{24.7\%} & \textbf{55.1\%} \\
			\midrule
			\multirow{2}{*}{4}&HSE & 20.5\% & 49.3\%  & 18.7\% & 48.1\%\\
			&Ours  & \textbf{24.9\%} & \textbf{55.9\%}& \textbf{24.8\%} & \textbf{55.3\%} \\
			\bottomrule
	\end{tabular}\vspace{-9pt}
	\end{center}
	\caption{Video/Paragraph retrival using N fixed-length segments on the text-truncated ActivityNet.}\label{tab:proposals}\vspace{-8pt}
	\end{table}
	%\smallskip
\begin{table}
\begin{center}
\begin{tabular}{lcc} 
\toprule
	Method & en$\rightarrow$zh&zh$\rightarrow$en\\
	\hline
	Vatex~\cite{wang2019vatex}&29.12\%&26.42\%\\
	Ours&\textbf{32.34\%}&\textbf{28.11\%} \\
				\bottomrule
\end{tabular}\vspace{-8pt}
\end{center}
\caption{Video-guided translation between English (en) and Chinese (zh). BLEU-4 scores.}\vspace{-17pt}\label{tab:vatex}
\end{table}
\subsection{Video-guided Machine Translation} \vspace{-2pt}
Finally, following~\cite{wang2019vatex}, we consider the task of  Video-guided Machine Translation (VMT), in which a sentence in a source language is translated to a target language, using both the source-text and a matching video. Similar to the video/paragraph retrieval task, our solution starts with training a retrieval model to embed the video and text closer together.  Since the videos are not segmented into clips, our model only utilizes the first hierarchy. We then fine-tune the video and sentence representations for the VMT, by using an LSTM decoder, followed by MLE loss on words. Like~\cite{wang2019vatex}, our LSTM has 1024 cells and a beam search with a width of 5 was used for evaluation. 
As is evident from the results, shown in Tab.~\ref{tab:vatex}, our method outperforms the method of~\cite{wang2019vatex}.\vspace{-6pt}

\section{Conclusions} \vspace{-4pt}
For retrieval, it is convenient to embed the data in each domain statically, which can be preprocessed and stored for future use. Nevertheless, our method highlights the importance of interactions or the cross-domain context when embedding the data of the two domains that are being matched. Meanwhile, if complexity is an issue for a specific application, one can consider using a shortlist that is obtained using static embeddings. In the future, we believe that combining conditioned embeddings and large-scale pretraining can lead to a significant performance gain.

\section*{Acknowledgments}
This project has received funding from the European Research Council (ERC) under the European Unions Horizon 2020 research, innovation programme (grant ERC CoG 725974), and  Grant No 2019783 from the United States-Israel Binational Science Foundation (BSF).

{\small
\bibliographystyle{ieee_fullname}
\bibliography{final_wacv}

\begin{thebibliography}{10}\itemsep=-1pt

\bibitem{aradhye2009video2text}
Hrishikesh Aradhye, George Toderici, and Jay Yagnik.
\newblock Video2text: Learning to annotate video content.
\newblock In {\em ICDM Workshops}, 2009.

\bibitem{ben2017mutan}
Hedi Ben-Younes, R{\'e}mi Cadene, Matthieu Cord, and Nicolas Thome.
\newblock Mutan: Multimodal tucker fusion for visual question answering.
\newblock In {\em ICCV}, 2017.

\bibitem{braude2021towards}
Tom Braude, Idan Schwartz, Alexander Schwing, and Ariel Shamir.
\newblock Towards coherent visual storytelling with ordered image attention.
\newblock {\em arXiv preprint arXiv:2108.02180}, 2021.

\bibitem{chen2019semantic}
Shaoxiang Chen and Yu-Gang Jiang.
\newblock Semantic proposal for activity localization in videos via sentence
  query.
\newblock In {\em AAAI}, 2019.

\bibitem{chen2020fine}
Shizhe Chen, Yida Zhao, Qin Jin, and Qi Wu.
\newblock Fine-grained video-text retrieval with hierarchical graph reasoning.
\newblock In {\em CVPR}, 2020.

\bibitem{cho2014properties}
Kyunghyun Cho, Bart Van~Merri{\"e}nboer, Dzmitry Bahdanau, and Yoshua Bengio.
\newblock On the properties of neural machine translation: Encoder-decoder
  approaches.
\newblock {\em SSST-8}, 2014.

\bibitem{dong2019dual}
Jianfeng Dong, Xirong Li, Chaoxi Xu, Shouling Ji, Yuan He, Gang Yang, and Xun
  Wang.
\newblock Dual encoding for zero-example video retrieval.
\newblock In {\em CVPR}, 2019.

\bibitem{duan2018weakly}
Xuguang Duan, Wenbing Huang, Chuang Gan, Jingdong Wang, Wenwu Zhu, and Junzhou
  Huang.
\newblock Weakly supervised dense event captioning in videos.
\newblock In {\em NIPS}, 2018.

\bibitem{fukui2016multimodal}
Akira Fukui, Dong~Huk Park, Daylen Yang, Anna Rohrbach, Trevor Darrell, and
  Marcus Rohrbach.
\newblock Multimodal compact bilinear pooling for visual question answering and
  visual grounding.
\newblock {\em EMNLP}, 2016.

\bibitem{gabeur2020multi}
Valentin Gabeur, Chen Sun, Karteek Alahari, and Cordelia Schmid.
\newblock Multi-modal transformer for video retrieval.
\newblock {\em ECCV}, 2020.

\bibitem{gao2017tall}
Jiyang Gao, Chen Sun, Zhenheng Yang, and Ram Nevatia.
\newblock Tall: Temporal activity localization via language query.
\newblock In {\em ICCV}, pages 5267--5275, 2017.

\bibitem{gao2019wslln}
Mingfei Gao, Larry~S Davis, Richard Socher, and Caiming Xiong.
\newblock Wslln: Weakly supervised natural language localization networks.
\newblock {\em EMNLP}, 2019.

\bibitem{anne2017localizing}
Lisa~Anne Hendricks, Oliver Wang, Eli Shechtman, Josef Sivic, Trevor Darrell,
  and Bryan Russell.
\newblock Localizing moments in video with natural language.
\newblock In {\em ICCV}, 2017.

\bibitem{huang2012multi}
Haiqi Huang, Yueming Lu, Fangwei Zhang, and Songlin Sun.
\newblock A multi-modal clustering method for web videos.
\newblock In {\em ISCTCS}, 2012.

\bibitem{kim2018bilinear}
Jin-Hwa Kim, Jaehyun Jun, and Byoung-Tak Zhang.
\newblock Bilinear attention networks.
\newblock In {\em NIPS}, 2018.

\bibitem{kim2016hadamard}
Jin-Hwa Kim, Kyoung-Woon On, Woosang Lim, Jeonghee Kim, Jung-Woo Ha, and
  Byoung-Tak Zhang.
\newblock Hadamard product for low-rank bilinear pooling.
\newblock {\em ICLR}, 2017.

\bibitem{krishna2017dense}
Ranjay Krishna, Kenji Hata, Frederic Ren, Li Fei-Fei, and Juan Carlos~Niebles.
\newblock Dense-captioning events in videos.
\newblock In {\em ICCV}, 2017.

\bibitem{lei2021less}
Jie Lei, Linjie Li, Luowei Zhou, Zhe Gan, Tamara~L Berg, Mohit Bansal, and
  Jingjing Liu.
\newblock Less is more: Clipbert for video-and-language learning via sparse
  sampling.
\newblock {\em CVPR}, 2021.

\bibitem{li2015hierarchical}
Jiwei Li, Minh-Thang Luong, and Dan Jurafsky.
\newblock A hierarchical neural autoencoder for paragraphs and documents.
\newblock {\em ACL}, 2015.

\bibitem{li2020hero}
Linjie Li, Yen-Chun Chen, Yu Cheng, Zhe Gan, Licheng Yu, and Jingjing Liu.
\newblock Hero: Hierarchical encoder for video+ language omni-representation
  pre-training.
\newblock {\em EMNLP}, 2020.

\bibitem{liu2019use}
Yang Liu, Samuel Albanie, Arsha Nagrani, and Andrew Zisserman.
\newblock Use what you have: Video retrieval using representations from
  collaborative experts.
\newblock {\em BMVC}, 2019.

\bibitem{lu2016hierarchical}
Jiasen Lu, Jianwei Yang, Dhruv Batra, and Devi Parikh.
\newblock Hierarchical question-image co-attention for visual question
  answering.
\newblock In {\em NIPS}, 2016.

\bibitem{luo2020univl}
Huaishao Luo, Lei Ji, Botian Shi, Haoyang Huang, Nan Duan, Tianrui Li, Jason
  Li, Taroon Bharti, and Ming Zhou.
\newblock Univl: A unified video and language pre-training model for multimodal
  understanding and generation.
\newblock {\em arXiv preprint arXiv:2002.06353}, 2020.

\bibitem{miech2019howto100m}
Antoine Miech, Dimitri Zhukov, Jean-Baptiste Alayrac, Makarand Tapaswi, Ivan
  Laptev, and Josef Sivic.
\newblock Howto100m: Learning a text-video embedding by watching hundred
  million narrated video clips.
\newblock In {\em ICCV}, 2019.

\bibitem{mithun2018learning}
Niluthpol~Chowdhury Mithun, Juncheng Li, Florian Metze, and Amit~K
  Roy-Chowdhury.
\newblock Learning joint embedding with multimodal cues for cross-modal
  video-text retrieval.
\newblock In {\em ICMR}, 2018.

\bibitem{mithun2019weakly}
Niluthpol~Chowdhury Mithun, Sujoy Paul, and Amit~K Roy-Chowdhury.
\newblock Weakly supervised video moment retrieval from text queries.
\newblock In {\em CVPR}, 2019.

\bibitem{niu2017hierarchical}
Zhenxing Niu, Mo Zhou, Le Wang, Xinbo Gao, and Gang Hua.
\newblock Hierarchical multimodal lstm for dense visual-semantic embedding.
\newblock In {\em ICCV}, 2017.

\bibitem{pan2016hierarchical}
Pingbo Pan, Zhongwen Xu, Yi Yang, Fei Wu, and Yueting Zhuang.
\newblock Hierarchical recurrent neural encoder for video representation with
  application to captioning.
\newblock In {\em CVPR}, 2016.

\bibitem{patrick2020support}
Mandela Patrick, Po-Yao Huang, Yuki Asano, Florian Metze, Alexander Hauptmann,
  Jo{\~a}o Henriques, and Andrea Vedaldi.
\newblock Support-set bottlenecks for video-text representation learning.
\newblock {\em ICLR}, 2020.

\bibitem{pennington2014glove}
Jeffrey Pennington, Richard Socher, and Christopher~D Manning.
\newblock Glove: Global vectors for word representation.
\newblock In {\em EMNLP}, 2014.

\bibitem{rohrbach2015dataset}
Anna Rohrbach, Marcus Rohrbach, Niket Tandon, and Bernt Schiele.
\newblock A dataset for movie description.
\newblock In {\em CVPR}, 2015.

\bibitem{schwartz2017high}
Idan Schwartz, Alexander~G Schwing, and Tamir Hazan.
\newblock High-order attention models for visual question answering.
\newblock {\em NIPS}, 2017.

\bibitem{schwartz2019simple}
Idan Schwartz, Alexander~G Schwing, and Tamir Hazan.
\newblock A simple baseline for audio-visual scene-aware dialog.
\newblock In {\em CVPR}, 2019.

\bibitem{schwartz2019factor}
Idan Schwartz, Seunghak Yu, Tamir Hazan, and Alexander~G Schwing.
\newblock Factor graph attention.
\newblock In {\em CVPR}, 2019.

\bibitem{tanLoGAN2019}
Reuben Tan, Huijuan Xu, Kate Saenko, and Bryan~A. Plummer.
\newblock Logan: Latent graph co-attention network for weakly-supervised video
  moment retrieval.
\newblock {\em WACV}, 2021.

\bibitem{venugopalan2015sequence}
Subhashini Venugopalan, Marcus Rohrbach, Jeffrey Donahue, Raymond Mooney,
  Trevor Darrell, and Kate Saenko.
\newblock Sequence to sequence-video to text.
\newblock In {\em ICCV}, 2015.

\bibitem{venugopalan2014translating}
Subhashini Venugopalan, Huijuan Xu, Jeff Donahue, Marcus Rohrbach, Raymond
  Mooney, and Kate Saenko.
\newblock Translating videos to natural language using deep recurrent neural
  networks.
\newblock {\em ACL}, 2015.

\bibitem{wang2019vatex}
Xin Wang, Jiawei Wu, Junkun Chen, Lei Li, Yuan-Fang Wang, and William~Yang
  Wang.
\newblock Vatex: A large-scale, high-quality multilingual dataset for
  video-and-language research.
\newblock In {\em ICCV}, 2019.

\bibitem{wei2009multimodal}
Shikui Wei, Yao Zhao, Zhenfeng Zhu, and Nan Liu.
\newblock Multimodal fusion for video search reranking.
\newblock {\em TKDE}, 2009.

\bibitem{xu2019multilevel}
Huijuan Xu, Kun He, Bryan~A Plummer, Leonid Sigal, Stan Sclaroff, and Kate
  Saenko.
\newblock Multilevel language and vision integration for text-to-clip
  retrieval.
\newblock In {\em AAAI}, 2019.

\bibitem{xu2016ask}
Huijuan Xu and Kate Saenko.
\newblock Ask, attend and answer: Exploring question-guided spatial attention
  for visual question answering.
\newblock In {\em ECCV}, 2016.

\bibitem{xu2016msr}
Jun Xu, Tao Mei, Ting Yao, and Yong Rui.
\newblock Msr-vtt: A large video description dataset for bridging video and
  language.
\newblock In {\em CVPR}, 2016.

\bibitem{xu2015show}
Kelvin Xu, Jimmy Ba, Ryan Kiros, Kyunghyun Cho, Aaron Courville, Ruslan
  Salakhudinov, Rich Zemel, and Yoshua Bengio.
\newblock Show, attend and tell: Neural image caption generation with visual
  attention.
\newblock In {\em ICML}, 2015.

\bibitem{yang2016stacked}
Zichao Yang, Xiaodong He, Jianfeng Gao, Li Deng, and Alex Smola.
\newblock Stacked attention networks for image question answering.
\newblock In {\em CVPR}, 2016.

\bibitem{yu2016video}
Haonan Yu, Jiang Wang, Zhiheng Huang, Yi Yang, and Wei Xu.
\newblock Video paragraph captioning using hierarchical recurrent neural
  networks.
\newblock In {\em CVPR}, 2016.

\bibitem{yu2018joint}
Youngjae Yu, Jongseok Kim, and Gunhee Kim.
\newblock A joint sequence fusion model for video question answering and
  retrieval.
\newblock In {\em ECCV}, 2018.

\bibitem{yuan2019find}
Yitian Yuan, Tao Mei, and Wenwu Zhu.
\newblock To find where you talk: Temporal sentence localization in video with
  attention based location regression.
\newblock In {\em AAAI}, 2019.

\bibitem{zhang2018cross}
Bowen Zhang, Hexiang Hu, and Fei Sha.
\newblock Cross-modal and hierarchical modeling of video and text.
\newblock In {\em ECCV}, 2018.

\bibitem{zhang2019man}
Da Zhang, Xiyang Dai, Xin Wang, Yuan-Fang Wang, and Larry~S Davis.
\newblock Man: Moment alignment network for natural language moment retrieval
  via iterative graph adjustment.
\newblock In {\em CVPR}, 2019.

\bibitem{ZhXuCoCVPR18}
Luowei Zhou, Chenliang Xu, and Jason~J Corso.
\newblock Towards automatic learning of procedures from web instructional
  videos.
\newblock In {\em AAAI}, 2018.

\bibitem{zhu2020actbert}
Linchao Zhu and Yi Yang.
\newblock Actbert: Learning global-local video-text representations.
\newblock In {\em CVPR}, 2020.

\end{thebibliography}
}

\end{document}